\documentclass[conference]{IEEEtran}
\usepackage{blindtext, graphicx, caption}
\usepackage[table,xcdraw]{xcolor}
\usepackage{booktabs}
\usepackage{multirow}
\usepackage{tabularx}
\ifCLASSINFOpdf
\else
\fi

\begin{document}
%
\title{Unconstrained Scene Text and Video Text Recognition for Arabic Script\vspace{-1em}}


\author{
\IEEEauthorblockN{Mohit Jain, Minesh Mathew and C. V. Jawahar}
\IEEEauthorblockA{Center for Visual Information Technology, IIIT Hyderabad, India.}
\IEEEauthorblockA{mohit.jain@research.iiit.ac.in, minesh.mathew@research.iiit.ac.in, jawahar@iiit.ac.in}
\vspace{0.7em}}



%


\maketitle
\thispagestyle{empty}

\begin{abstract}
Building robust recognizers for Arabic has always been challenging.
We demonstrate the effectiveness of an end-to-end trainable \textsc{cnn-rnn} hybrid architecture in recognizing Arabic text in videos and natural scenes. We outperform previous state-of-the-art on two publicly available video text datasets - \textsc{alif} and \textsc{activ}. For the scene text recognition task, we introduce a new Arabic scene text dataset and establish baseline results. For scripts like Arabic, a major challenge in developing robust recognizers is the lack of large quantity of annotated data. We overcome this by synthesizing millions of  Arabic text images from a large vocabulary of Arabic words and phrases.
Our implementation is built on top of the model introduced here~\cite{shi}  which is proven quite effective for English scene text recognition. The model follows a segmentation-free, sequence to sequence transcription approach. The network transcribes a sequence of convolutional features from the input image to a sequence of target labels. This does away with the need for segmenting input image into constituent characters/glyphs, which is often difficult for Arabic script. Further, the ability of \textsc{rnn}s to model contextual dependencies yields superior recognition results.  
 
\end{abstract}

\begin{IEEEkeywords}
Arabic, Arabic Scene Text, Arabic Video Text, Synthetic Data, Deep Learning, Text Recognition. 
\end{IEEEkeywords}

%
\IEEEpeerreviewmaketitle

\section{Introduction}
For many years, the focus of research on text recognition in Arabic has been on printed and handwritten documents~\cite{prasad, hajj, amin96, amin98}. Majority of the works in this space were on individual character recognition~\cite{amin96, amin98}. 
Since segmenting an Arabic word or line image into its sub units is challenging, such models did not scale well.
In recent years there has been a shift towards segmentation-free text recognition models, mostly based on Hidden Markov Models (\textsc{hmm}) or Recurrent Neural Networks (\textsc{rnn}). Such models generally follow a sequence-to-sequence approach wherein the input line/word image is directly transcribed to a sequence of labels.
Methods such as \cite{yousefi} and \cite{hasan} use \textsc{rnn} based models for recognizing printed/handwritten Arabic script.
Our attempt to recognize Arabic text on videos (video text recognition) and natural scenes (scene text recognition) follows the same paradigm.
Video text recognition in Arabic has gained interest recently~\cite {yousfi,activ,halima}. There are now two benchmarking datasets available for the task \textsc{alif}~\cite{yousfi} and \textsc{activ}~\cite{activ}.
In the scene text domain, \cite{tounsi} uses sparse coding of SIFT features for a bag-of-features model using spatial pyramid matching for the recognition task at the character level in Arabic. To the best of our knowledge, there has been no work so far on scene text recognition at word level in Arabic.

\begin{figure}[htp]
\centering
\captionsetup{justification=centering}
\includegraphics[width=\columnwidth, height=150pt]{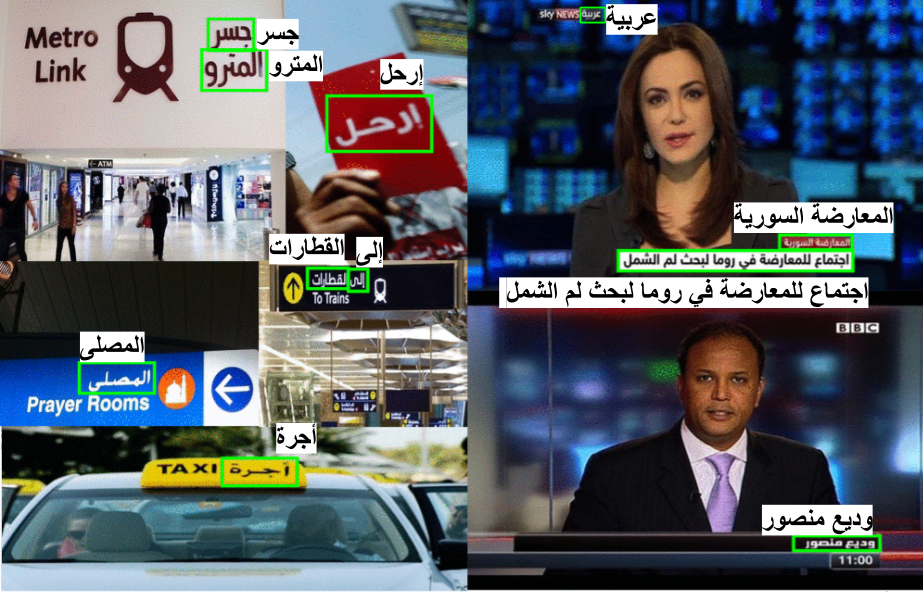}
\caption{Examples of Arabic Scene Text (left)  and  Video Text (right) recognized by our model. Our work deals only with the recognition of cropped words/lines. The bounding boxes were provided manually.}
\label{fig:fig6}
\vspace{-0.5cm}
\end{figure}

The computer vision community experienced a strong revival of neural networks based solutions in the form of Deep Learning in recent years. This process was stimulated by the success of models like Deep Convolutional Neural Networks (\textsc{dcnn}s) in feature extraction, object detection and classification tasks as seen in~\cite{krizhevsky,girshick}. These tasks however pertain to a set of problems where the subjects appear in isolation, rather than appearing as a sequence. Recognising a sequence of objects often requires the model to predict a series of description labels instead of a single label. \textsc{dcnn}s are not well suited for such tasks as they generally work well in scenarios where the inputs and outputs are bounded by fixed dimensions. Also, the lengths of these sequence-like objects might vary drastically and that escalates the problem difficulty. Recurrent neural networks (\textsc{rnn}s) tackle the problems faced by \textsc{dcnn}s for sequence-based learning by performing a forward pass for each segment of the sequence-like-input. 
Such models often involve a pre-processing/feature-extraction step, where the input is first converted into a sequence of feature vectors~\cite{graves09, su}.
These feature extraction stage would be independent of \textsc{rnn}-pipeline and hence they are not end-to-end trainable.

The upsurge in video sharing on social networking websites and the increasing number of TV channels in today's world reveals videos to be a fundamental source of information. Effectively managing, storing and retrieving such video data is not a trivial task. Video text recognition can greatly aid video content analysis and understanding, with the recognized text giving a direct and concise description of the stories being depicted in the videos. In news videos, superimposed tickers running on the edges of the video frame are generally highly correlated to the people involved or the story being portrayed and hence provide a brief summary of the news event. 

Reading text in natural scenes is relatively harder task compared to printed  text recognition. The problem has been drawing increasing research interest in recent years. This can partially be attributed to the rapid development of wearable and mobile devices such as smart phones, google glass and self-driving cars, where scene text is a key module to a wide range of practical and useful applications. While the recognition of text in printed  documents has been studied extensively in the form of Optical  Character Recognition (\textsc{ocr}) Systems, these methods generally do not generalize very well to a natural scene setting where factors like inconsistent lighting conditions, variable fonts, orientations, background noise and image distortions add to the problem complexity. Even though Arabic is the fourth most spoken language in the world after Chinese, English and Spanish, to the best of our knowledge, there has not been any work in the field of word-level scene text recognition, which makes this area of research ripe for exploration.

\subsection{Intricacies of Arabic Script}
Automatic recognition of Arabic is a pretty challenging task due to various intricate properties of the script. There are only 28 letters in the script and it is written in a semi-cursive fashion from right to left. The letters of the script are generally connected by a line at its bottom to the letters preceding and following it, except for 6 letters which can be linked only to their preceding letters. Such an instance in a word creates a \textit{paw} (part-of Arabic word) in the word. Arabic script has another exceptional characteristic where the shape of a letter changes depending on whether the letter appears in the word in isolation, at the beginning, middle or at the end. In general, each letter can have one to four possible shapes which might have little or no similarity in shape whatsoever. Another frequent occurrence in Arabic is that of dots. The addition of a single dot to a letter can change the character completely. Arabic also follows ligature, where two letters when combined form a third different letter in a way that they cannot be separated by a simple base-line (i.e. a complex shape represents this combined letter). These distinctive characteristics make automated Arabic script recognition more challenging than most other scripts. 
 
\subsection{Scene Text and Video Text Recognition}
Though there has been a lot of work done in the field of text transcription in natural scenes and videos for the English script~\cite{graves09, su, jaderberg, wang}, it is still in a nascent state as far as Arabic script is considered. Previous attempts made to address similar problems in English~\cite{wang, bissacco} first detect individual characters and then character-specific \textsc{dcnn} models are used to recognize these detected characters. The short-comings of such methods are that they require training a strong character detector for accurately detecting and cropping each character out from the original word. In case of Arabic, this task becomes even more difficult due to the intricacies of the script, as discussed earlier. Another approach by Jaderberg et al.~\cite{jaderberg}, was to treat scene text recognition as an image classification problem instead. To each image, they assign a class label from a lexicon of words spanning the English language (90K most frequent words were chosen to put a bound on the span-set). However, this approach is limited to the size of lexicon used for its possible unique transcriptions and the large number of output classes add to training complexity. Hence the model is not scalable to inflectional languages like Arabic where number of unique words is much higher compared to English. Another category of solutions typically embed image and its text label in a common subspace and retrieval/recognition is performed on the learnt common representations. For example, Almazan et al.~\cite{almazan} embed word images and text strings in a common vector-subspace, and thus convert the task of word recognition into a retrieval problem. Yao et al.~\cite{yao} and Gordo et al.~\cite{gordo} used mid-level features for scene text recognition.

The segmentation-free, transcription approach was proved quite effective for Indian Scripts \textsc{ocr}~\cite{naveen, minesh} where segmentation is often problematic. Similar approach was used in~\cite{su} for scene text recognition. Handcrafted features derived from image gradients were used with a \textsc{rnn} to map the sequence of features to a sequence of labels. Unlike the problem of \textsc{ocr}, scene text recognition required more robust features to yield results comparable to the transcription based solutions for \textsc{ocr}.
 A novel approach combining the robust convolutonal features and transcription abilities of \textsc{rnn} was introduced by~\cite{shi}. We employ this particular architecture  where the hybrid \textsc{cnn-rnn} network with a Connectionist Temporal Classification (\textsc{ctc}) loss is trained end-to-end.

Works on text extraction from videos has generally been in four broad categories, edge detection methods~\cite{agnihotri99, agnihotri02, hua}, extraction using connected components~\cite{jain, lienhart}, texture classification methods~\cite{karray} and correlation based methods~\cite{karray05, kherallah}. The previous attempts at solving video text for Arabic used two separate routines; one for extracting relevant image features and another for classifying features to script labels for obtaining the target transcription. In~\cite{halima} text-segmentation and statistical feature extraction are followed by fuzzy k-nearest neighbour techniques to obtain transcriptions. Similarly, \cite{alif} experiment with two feature extraction routines; \textsc{cnn} based feature extraction and Deep Belief Network (\textsc{dbn}) based feature extraction, followed by a Bi-directional Long Short Term Memory (\textsc{lstm}) layer~\cite{gers, hochreiter}.

The rest of the paper is organized as follows; Section II describes  the hybrid \textsc{cnn-rnn} architecture we use. Section III focuses on the Arabic text recognition from a transcription perspective. Section IV presents the results on benchmarking datasets, followed by a discussion based on the qualitative results. Section V concludes with the findings of our work.

\section{Hybrid \textsc{cnn-rnn} Architecture}
The network consists of three components. The initial convolutional layers, the middle recurrent layers and a final transcription layer. This architecture we employ was introduced in~\cite{shi}.

\begin{figure*}[!htp]
\centering
\captionsetup{justification=centering}
\includegraphics[width=\textwidth, height=4cm]{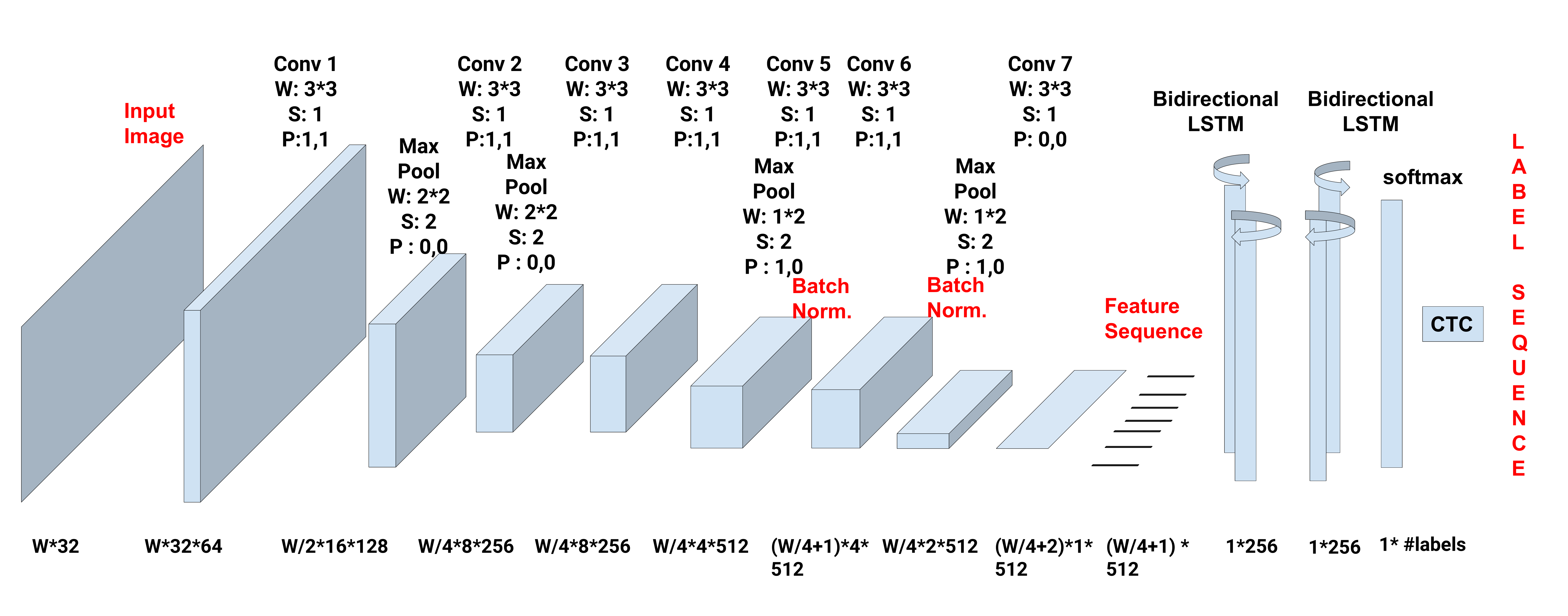}
\caption{The deep neural network architecture employed in our work, which was devised by~\cite{shi}. \\The symbols `k', `s' and `p' stand for kernel size, stride and padding size respectively.}
\label{fig:fig3}
\vspace{-1em}
\end{figure*}

 The role of the convolutional layers is to extract robust feature representations from the input image and feed it into the recurrent layers, which in turn will transcribe them to an output sequence of labels or characters. The sequence to sequence transcription is achieved by a \textsc{ctc} layer at the output. 

All the images are scaled to a fixed height before being fed to the convolutional layers. The convolutional components then create a sequence of feature vectors from the feature maps by splitting them column-wise, which then act as inputs to the recurrent layers (Fig.~\ref{fig:fig3}).  
This feature descriptor is highly robust and most importantly can be trained to be adopted to a wide variety of problems~\cite{krizhevsky, girshick, jaderberg}.

On top of the convolutional layers, the recurrent layers take each frame from the feature sequence generated by the convolutional layers and make predictions. The recurrent layers consist of deep bidirectional \textsc{lstm} nets. 
Number of parameters in a \textsc{rnn} is independent of the length of the sequence, hence we can simply unroll the network as many times as the number of time-steps in the input sequence. This in our case helps perform unconstrained recognition as the predicted output can be any sequence of labels derived from the entire label set.
Traditional \textsc{rnn} units \textit{(vanilla \textsc{rnn}s)} face the problem of vanishing gradients~\cite{bengio} and hence (\textsc{lstm}) units are used which were specifically designed to tackle the vanishing-gradients problem~\cite{gers, hochreiter}. 
In a text recognition problem, contexts from both directions (left-to-right and right-to-left) are useful and complementary to each other in outputting the right transcription. Therefore, a combined forward and backward oriented \textsc{lstm} is used to create a bi-directional \textsc{lstm} unit. Multiple such bi-directional \textsc{lstm} layers can be stacked to make the network deeper and gain higher levels of abstractions over the image-sequences as shown in~\cite{graves13}.

The transcription layer at the top of the network is used to translate the predictions generated by the recurrent layers into label sequences for the target language. The \textsc{ctc} layer's conditional probability is used in the objective function as shown in~\cite{graves06}. The complete network configuration used for the experiments can be seen in (Fig.~\ref{fig:fig3}).

\section{Arabic Text Transcription}
The efficacy of a seq2seq approach lies in the fact that the input sequence can be transcribed directly to the output sequence without the need for a target defined at each time-step. This was proven quite effective in recognizing complex scripts where sub-word segmentation is often difficult~\cite{minesh}. In addition, contextual modelling of the sequence in both the directions, forward and backward, is achieved by the bidirectional \textsc{lstm} block. Contextual information is critical in making accurate predictions for a Script like Arabic where there are many similar looking characters/glyphs. The hybrid architecture replaces the need for any handcrafted features to be extracted from the image. Instead, more robust convolutional features are fed to the \textsc{rnn} layers. 


\begin{figure}[htp]
\centering
\captionsetup{justification=centering}
\includegraphics[width=\columnwidth, height=70pt]{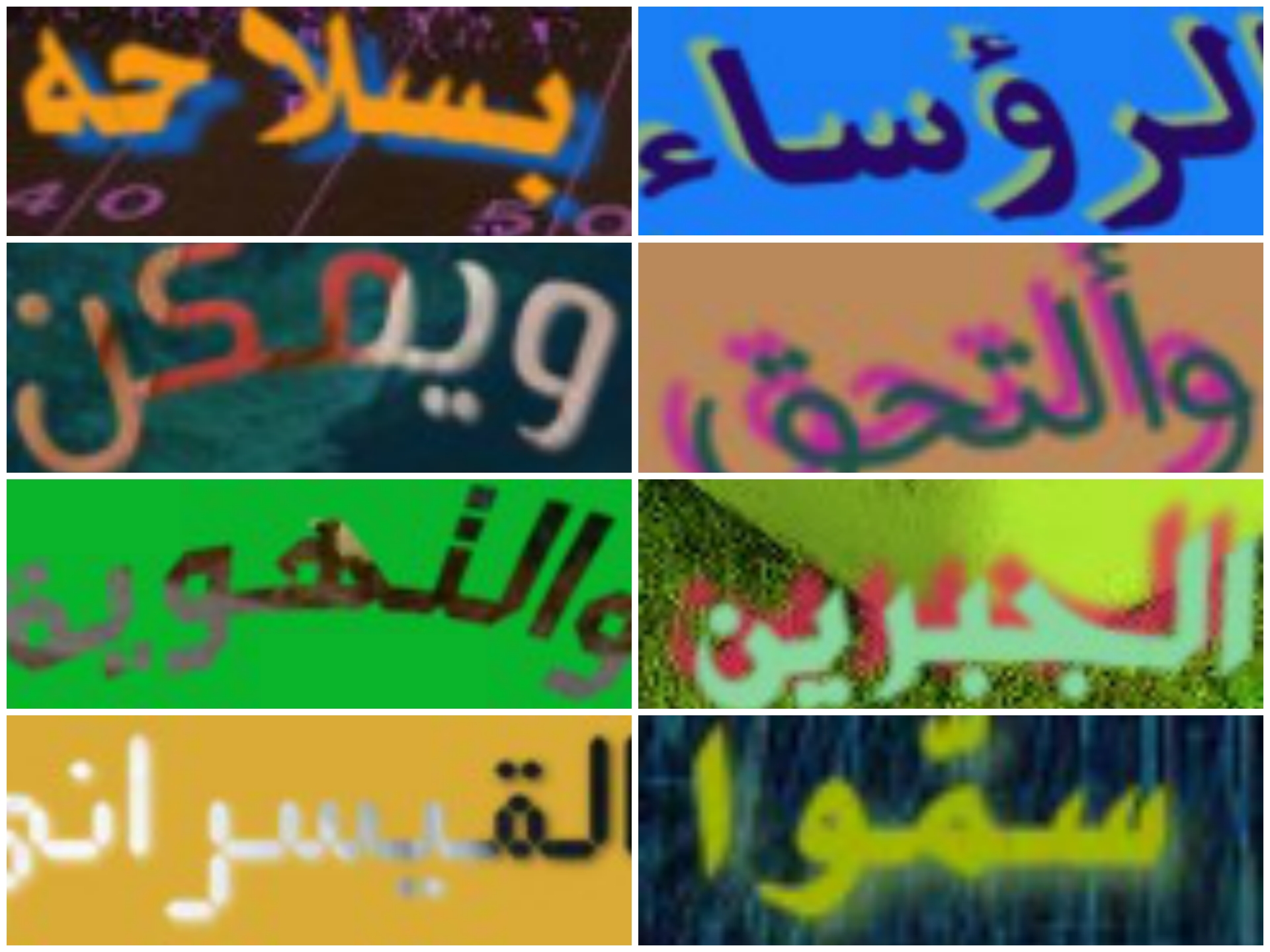}
\caption{Sample images from the rendered synthetic Arabic scene text dataset. The images closely resemble real world scene images.}
\label{fig:fig2}
\vspace{-0.5cm}
\end{figure}

\subsection{Datasets}
Models for both video text and scene text recognition problems are trained using synthetic images rendered from a large vocabulary using freely available Arabic Unicode fonts (Fig.~\ref{fig:fig2}). For scene text, the images were rendered from a vocabulary of a quarter million most commonly used words in Arabic Wikipedia. A random word from the vocabulary is first rendered into the foreground layer of the image by varying the font, stroke color, stroke thickness, kerning, skew and rotation. Later a random perspective projective transformation is applied to the image, which is then blended with a random crop from a natural scene image. Finally the foreground layer is alpha composed with a background layer, which is again a random crop from a random natural image. The synthetic line images for video text recognition are rendered from random text lines crawled from Arabic news websites. The rendering process is much simpler since real video text usually have uniform color for the text and background and lacks any perspective distortion.  
Details on the rendering process are described here~\cite{iiit-synth}. Around 2 million video text line images and 4 million scene text word images were used for training the respective models. The model for video text recognition task was initially trained on the Synthetic video text dataset and then fine-tuned on the train partitions of real-world datasets, \textsc{alif} and \textsc{activ}.

\textsc{alif} dataset consists of 6,532 cropped text line images from 5 popular Arabic News channels. \textsc{activ} dataset is larger than \textsc{alif} and contains 21,520 line images from popular Arabic News channels. The dataset contains video frames wherein bounding boxes of text lines are annotated. 

\begin{figure}[htp]
\centering
\captionsetup{justification=centering}
\includegraphics[width=\columnwidth]{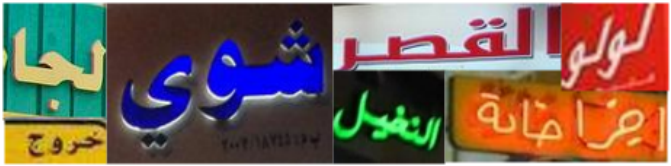}
\caption{Sample images from the Arabic scenetext dataset.}
\label{fig:fig4}
\vspace{-2em}
\end{figure}

A new Arabic scene text  dataset was curated by downloading freely available images containing Arabic script from Google Images. The dataset consists of 2000 word images of Arabic script occurring in various scenarios like local markets \& shops, billboards, navigation signs, graffiti, etc, and spans a large variety of naturally occurring image-noises and distortions (Fig.~\ref{fig:fig4}). The images were manually annotated by human experts of the script and the transcriptions as well as the image data is being made publicly available for future research groups to compare and improve performance upon.

\subsection{Implementation Details}
Convolutional block in the network follows the \textsc{vgg} architecture~\cite{simonyan}. In the 3rd and 4th max-pooling layers, the pooling windows used are rectangular instead of the usual square windows used in \textsc{vgg}. The advantage of doing this is that the feature maps obtained after the convolutional layers are wider and hence we obtain longer feature sequences as inputs for the recurrent layers that follow. To enable faster batch learning, all input images are re-sized to a fixed width and height (32x100 for scene text and 32x504 for video text). We have observed that resizing all images to a fixed width is not affecting the performance much. The images are horizontally flipped before feeding them to the convolutional layers since Arabic is read from right to left. 

To tackle the problems of training such deep convolutional and recurrent layers, we used the \textit{batch normalization}~\cite{ioffe} technique. 
The network is trained with stochastic gradient descent (\textsc{sgd}). Gradients are calculated by the back-propagation algorithm. Precisely, the transcription layers' error differentials are back-propagated with the forward-backward algorithm, as shown in~\cite{graves06}. While in the recurrent layers, the Back-Propagation Through Time (\textsc{bptt})~\cite{werbos} algorithm is applied to calculate the error differentials. The hassle of manually setting the learning-rate parameter is taken care of by using \textsc{adadelta} optimization~\cite{zeiler}. 

\begin{figure*}[htp]
\centering
\captionsetup{justification=centering}
\includegraphics[width=\textwidth, height=90pt]{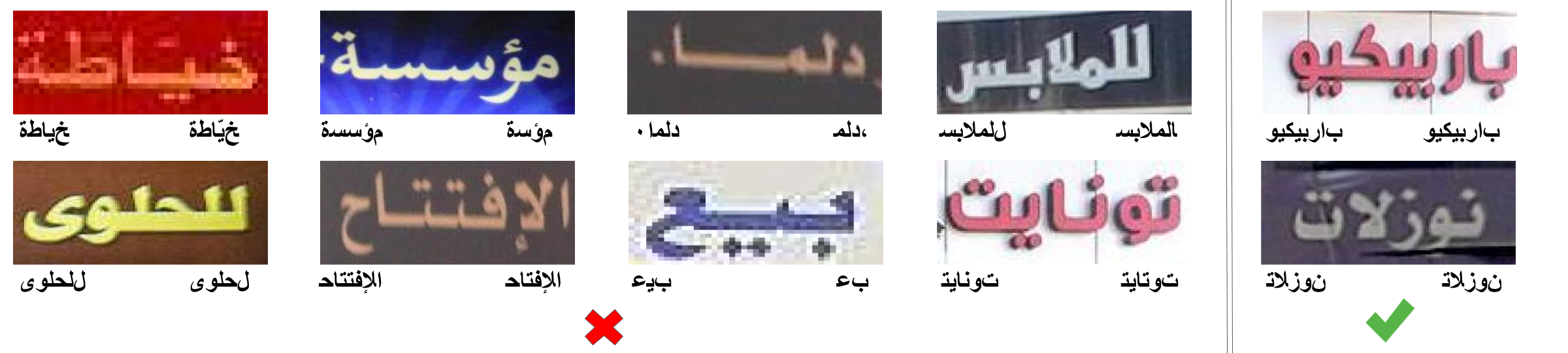}
\caption{Qualitative results of Scene Text Recognition. For each image, the annotations on bottom-left and bottom-right are the label and model prediction, respectively.}
\label{fig:fig5}
\vspace{-2em}
\end{figure*}
\section{Results and Discussion}
In this section we demonstrate the efficacy of above architecture in recognizing Arabic script appearing in video frames and natural scene images. It is assumed that input images are cropped word images from natural scenes, not full scene images. Since there were no previous works in Arabic scene text recognition at word level, the baseline results are reported on the new Arabic scene text we introduce. For video text recognition problem the results are reported on two existing video text datasets -  \textsc{alif}~\cite{alif} and \textsc{activ}~\cite{activ}.

Results on video text recognition are presented in Table I.
Since there has been no work done in word-level Arabic scene text recognition, we compare the results obtained on the Arabic scene text dataset using a popular free English \textsc{ocr} - Tesseract~\cite{tesseract}.
The performance has been evaluated using the following metrics; \textit{\textsc{crr} - Character Recognition Rate, \textsc{wrr} - Word Recognition Rate, \textsc{lrr} - Line Recognition Rate}. In the below equations, \textit{RT} and \textit{GT} stand for recognized text and ground truth respectively.

{\fontsize{8}{4}\selectfont 
\[CRR = \frac{(nCharacters - \sum EditDistance(RT, GT))}{nCharacters}\]
\vspace{-0.3em}
\[WRR = \frac{nWordsCorrectlyRecognized}{nWords}\]
\vspace{-0.3em}
\[LRR = \frac{nTextImagesCorrectlyRecognized}{nImages}\]
}

\vspace{-1em}
It should be noted that even though the methods compared on for video text recognition use a separate convolutional architecture for feature extraction, unlike the  end-to-end trainable  architecture, we obtain better character and line-level accuracies for the Arabic video text recognition task and set the new state-of-the-art for the same.

\vspace{-1em}

\begin{table}[h!]
\centering
\captionsetup{justification=centering}
\caption{Accuracy for Video Text Recognition.}
\vspace{-0.5em}
\resizebox{\columnwidth}{!}{\begin{tabular}{|c|cc|cc|cc|}
\hline
\multicolumn{1}{|c|}{\cellcolor[HTML]{C0C0C0}\textit{VideoText}} & \multicolumn{2}{c|}{\textbf{ALIF Test1}} & \multicolumn{2}{c|}{\textbf{ALIF Test2}} & \multicolumn{2}{c|}{\textbf{AcTiV}}   \\ \cline{2-7} 
\cellcolor[HTML]{C0C0C0}                                         & \textit{CRR(\%)}    & \textit{LRR(\%)}   & \textit{CRR(\%)}    & \textit{LRR(\%)}   & \textit{CRR(\%)}   & \textit{LRR(\%)} \\ \hline
ConvNet-BLSTM \cite{alif}                                    & 94.36                  & 55.03              & 90.71                  & 44.90              & --  & --             \\
DBN-BLSTM \cite{alif}                                        & 90.73                  & 39.39              & 87.64                  & 31.54              & --  & --             \\
ABBYY \cite{abbyy}                                            & 83.26                  & 26.91              & 81.51                  & 27.03              & -- & --              \\ \hline
\textbf{\textsc{cnn-rnn} hybrid network}                                               & \textbf{98.17}                  & \textbf{79.67}        & \textbf{97.84}                  & \textbf{77.98}                 & \textbf{97.44}      & \textbf{67.08}               \\ \hline
\end{tabular}}
\vspace{-1em}
\caption*{\small{The hybrid \textsc{cnn-rnn} architecture we employ outperforms previous video text recognition benchmarks.}}
\label{my-label}
\end{table}

\vspace{-2em}

\begin{table}[h!]
\centering
\captionsetup{justification=centering, margin=2em}
\caption{Accuracy for Scene Text Recognition.}
\label{my-label1}
\vspace{-0.5em}
\resizebox{0.7\columnwidth}{!}{\begin{tabular}{|c|cc|}
\hline
\cellcolor[HTML]{C0C0C0}\textit{SceneText} & \multicolumn{2}{c|}{\textbf{Arabic SceneText Dataset}} \\ \cline{2-3} 
\cellcolor[HTML]{C0C0C0}                   & \textit{CRR(\%)}    & \textit{WRR(\%)}    \\ \hline
Tesseract \cite{tesseract}                              & 17.07               & 5.92      \\ \hline
\textbf{Baseline using the hybrid \textsc{cnn-rnn}}                              & \textbf{75.05}               & \textbf{39.43}      \\ \hline
\end{tabular}}
\vspace{-1em}
\caption*{\small{Lower accuracies on scene text recognition problem testify the inherent difficulty associated with the problem compared to printed or video text recognition.}}
\vspace{-1em}
\end{table}

\vspace{-1em}
 
On video text recognition, we report the best results so far on both the datasets. Scene text recognition is a much harder problem compared to \textsc{ocr} or video text recognition. The variability in terms of lighting, distortions and typography make the learning pretty hard and we provide baseline results for the same. A qualitative analysis of the recognition tasks can be seen in Fig.~\ref{fig:fig5}.

\section{Conclusion}
We demonstrate that the state-of-the-art deep learning techniques can be successfully adapted to some rather challenging tasks like Arabic text recognition. The newer script and language agnostic approaches are well suited for low resource languages like Arabic where the traditional methods often involved language specific modules. The success of \textsc{rnn}s in sequence learning problems has been instrumental in the recent advances in speech recognition and image to text transcription problems. This came as a boon for languages like Arabic where the segmentation of words into sub word units was often troublesome. The sequence learning approach could directly transcribe the images and also model the context in both forward and backward directions. With better feature representations and learning algorithms available, we believe the focus should now shift to harder problems like scene text recognition. We hope the introduction of a new Arabic scene text dataset and the initial results would instill an interest among the Arabic computer vision community to pursue this field of research further.


%


\vspace{-0.3em}

\section*{Acknowledgment}
\vspace{-0.3em}
The authors would like to thank Maaz Anwar, Anjali, Saumya, Vignesh and Rohan for helping annotate the Arabic scenetext dataset and Dr. Girish Varma for his timely help and discussions.

\vspace{-1.7em}

\ifCLASSOPTIONcaptionsoff
  \newpage
\fi



%

%

\begin{IEEEbiography}[{\includegraphics[width=1in,height=1.25in,clip,keepaspectratio]{picture}}]{John Doe}
\blindtext
\end{IEEEbiography}




\end{document}